# How Do Drivers Behave at Roundabouts in a Mixed Traffic? A Case Study Using Machine Learning

Farah Abu Hamad[a], Rama Hasiba[a], Deema Shahwan[a], and Huthaifa I. Ashqar[b,c]

[a]*An-Najah National University, Nablus, Palestine.*
[b]*Precision Systems, Inc., Washington, DC, 20003, United States*
[c]*Arab American University, P.O. Box 240, 13 Zababde, Jenin, Palestine.*

**Abstract**

Driving behavior is considered a unique driving habit of each driver and has a significant impact on road safety. Classifying driving behavior and introducing policies based on the results can reduce the severity of crashes on the road. Roundabouts are particularly interesting because of the interconnected interaction between different road users at the area of roundabouts, which different driving behavior is hypothesized. This study investigates driving behavior at roundabouts in a mixed traffic environment using a data-driven unsupervised machine learning to classify driving behavior at three roundabouts in Germany. We used a dataset of vehicle kinematics to a group of different vehicles and vulnerable road users (VRUs) at roundabouts and classified them into three categories (i.e., conservative, normal, and aggressive). Results showed that most of the drivers proceeding through a roundabout can be mostly classified into two driving styles: conservative and normal because traffic speeds in roundabouts are relatively lower than in other signalized and unsignalized intersections. Results also showed that about 77% of drivers who interacted with pedestrians or cyclists were classified as conservative drivers compared to about 42% of conservative drivers that did not interact or about 51% from all drivers. It seems that drivers tend to behave abnormally as they interact with VRUs at roundabouts, which increases the risk of crashes when an intersection is multimodal. Results of this study could be helpful in improving the safety of roads by allowing policymakers to determine the effective and suitable safety countermeasures. Results will also be beneficial for the Advanced Driver Assistance System (ADAS) as the technology is being deployed in a mixed traffic environment.

## 1. Introduction

Driver behavior describes the actions that drivers take while driving at different road infrastructures. In the last decades, many researchers tend to investigate drivers' behavior as it become challenging due to the dramatically increasing of vehicle users and interactions with other road users, specifically, pedestrians and bicycles. Understanding the behaviors of drivers could lead to enhance the applicability of advanced driving assistants (ADS), safer roads by reducing crash severity, and increase the effectiveness of using Intelligent transportation system (ITS) for traffic operation. In addition, we might better understand the relationship between driving behavior and the annual increase in road crashes. As plenty of factors can affect such behaviors like age, gender, emotions, and experience, it is also different for the same driver in different situations including different road infrastructures. Various methods were used to investigate driving behavior. According to a study in [1], [2], data collection methods include surveys, questionnaires, simulations, roadside camera observations, and naturalistic experiments. It seems that working on data that have been extracted from naturalistic experiments is suitable due to the high percentage of reliability. From that perspective, many datasets on driver behavior have been collected for that goal, such as HDD [3], highD [4], inD [5], rounD [6]. Classifying driving behavior is also useful in understanding and mimicking humans' behavior so it can be used for safety and operation purposes. One of the effective ways to classify is using classification algorithms in machine learning which includes unsupervised algorithms such as K-means, and supervised such as Decision Trees (DL), Naïve Bayes, Support vector machines (SVMs), and deep learning [7].

Understanding driving behavior at roundabouts is important for several reasons [8]–[12]. First, roundabouts are designed to reduce the severity and frequency of accidents compared to traditional intersections, but the safety benefits depend on drivers following the correct behavior. Understanding how drivers behave at roundabouts can help identify potential safety hazards and inform improvements to the design and operation of roundabouts. Second, roundabouts can improve traffic flow by reducing delays and minimizing the need for traffic signals or stop signs. However, traffic flow can be affected by driver behavior, such as improper lane use or failure to yield to other vehicles. Understanding driving behavior at roundabouts can help identify areas where traffic flow can be improved. Third, efficient use of roundabouts depends on drivers using proper behavior, such as entering and exiting the roundabout in the correct lane, yielding to other vehicles already in the roundabout, and signaling their intentions. Fourth, compliance with traffic rules and regulations is essential for the safe and efficient operation of roundabouts. Understanding driving behavior at roundabouts can help identify areas where drivers are not complying with traffic rules, which can inform enforcement efforts and education campaigns.

In this study, we used a dataset from three roundabouts in Germany to classify driving behavior into three styles, namely, conservative, normal, and aggressive, using an unsupervised machine learning for clustering. We also investigated the driving behavior of drivers who interacted with pedestrians and bicycles (VRUs) going through the roundabouts. We compared the resulted behavior of interacted drivers with pedestrians and bicycles and drivers who did not interact with them. To the best of our knowledge, this is the first study in the literature that addresses the two abovementioned contributions.

## 2. Literature Review

Understanding the behavior of road users is of vital importance for the development of trajectory prediction systems. Moreover, for a successful market launch of automated vehicles (AVs), proof of their safety is essential. While there has been much research on several datasets and different types of trajectories of road users, bunch of researchers have taken the roundabouts into consideration as it describes a high level of complexity. Measurement data should be collected at a reasonable effort, contain naturalistic behavior of road users, and include all data relevant for a description of the identified scenarios in sufficient quality. Human drivers naturally use their knowledge of other road users' behaviors to improve their driving and the safety of the traffic. A study considered two three-leg junctions and one roundabout to understand how at grade intersections affect driving behavior by comparing the drivers' stress levels using Electrodermal activity. The stress level induced by each type of intersection was evaluated through an Electrodermal Impact Index (EEI). Results suggested that the stress level induced by roundabouts is more than double that induced by standard intersections [13]. Another study used data from five roundabouts in addition to a questionnaire that has been randomly distributed to drivers to explore driving behavior. Results showed that the percentage of drivers breaching at least one traffic regulation is approximately 90% of all drivers. Leaving without flashing and entering the roundabout without giving way were the most frequent violation types [9]. A similar study distributed a questionnaire to obtain the needed information then linking them with the real situation on roundabouts. Analyzing the data showed a large percent of drivers have a good knowledge regarding roundabout rules. A few modifications were done on two roundabouts to compare between before and after based on measures of effectiveness. The analyzed data showed that for vital areas and for traffic volumes greater than 3000 veh/h; the level of service ranges between B and C, and the control delay ranges between 10 s to 30 s. The study helped traffic planners and designers in the decision-making process providing several intersection alternatives between roundabouts and signalized intersection, where the impact of driving behavior should be considered [10].

Detecting risk driving and the prediction of drivers' behavior intentions is necessary to maintain the safety of road users and raise the success rate of driverless vehicles. Many studies have investigated the nature and variation of driving risk in roundabouts, to allow connected vehicles to quickly assess a personalized and real-time level of risk associated with crossing a roundabout. One study recorded time to collision (TTC) at roundabouts, then applied machine learning on the data to assess the probability that a vehicle will choose the upcoming exit. A risk metric was developed based on the TTC data and the probability. The results show a strong relation with the coefficient of variation of TTC values on roundabouts. The obtained risk knowledge has the potential to support driver assistance systems in roundabouts [12]. Another study took into consideration the steering wheel angle, angle velocity, and vehicle position to predict whether the driver will take the upcoming exit or not. They collected data of driving behaviors to model human driving behavior in interaction with roundabouts by using support vector machine – a supervised machine learning model that analyze data for classification and regression analysis. From the experimental results, the vehicles position can be estimated in which the prediction becomes reliable [8]. A study presented two methods to estimate when the driver leaves a roundabout based on the behavior of the drivers. The first method starts with training data to extract typical behavior patterns, then using it to classify other drivers' intentions. The second method does not require a training data. It generates the typical behavior patterns from a precise map and the classification was done on arbitrary roundabouts if the map is available. Results showed that the performance of the map-based approach is comparable to the data-driven approach [14].

Many dynamic factors influence drivers' behaviors such as speed, acceleration, circulating flow of the potentially conflicting vehicles. A study analyzed these factors in addition of driving behavior characteristics then applied Random Forest algorithm to predict the driving behavior. Using a simulator to mimicking real driving conditions using traffic participants with different motion styles, four typical roundabouts were created to collect the data. Random forest model has good performance in predicting the roundabout behaviors of human drivers. Results show that the geometric parameters have little contribution for predicting the driving behavior. The relative velocity between surrounding vehicle and master vehicle are most contributively to human driving behavior [11]. Another study focused on the behavior of drivers at turbo-roundabouts as well as the kinematic parameters of the vehicle (i.e., speed and acceleration). The calibration of traffic microsimulation models or the assignment of behavior parameters to closed-form capacity models can both benefit from empirical evaluations of these parameters. The study's findings revealed that vehicle speeds in entry lanes are quite low (below 25 km/h, 15 m before the yield line), and that ring lane accelerations typically have values below 1.5 m/s$^2$ [15].

## 3. Dataset and Methods

**Dataset**

The aim of this study is to develop a classification model to classify drivers on road user trajectories using rounD dataset. The rounD dataset is a dataset of naturalistic road user trajectories recorded at three roundabouts in Germany. The dataset is applicable on many tasks such as road user prediction, driver modelling, scenario-based safety validation of automated driving systems. The dataset includes vehicles of different classes (i.e., car, truck, trailer, van, bus), pedestrians, bicyclists, and motorcycles. In addition, data is collected for a total of six hours of recording with more than 13,746 road users. Furthermore, the dataset was held at three different recording locations on different roundabout types with typical positioning error <10 cm as shown in Fig. 1 [6].

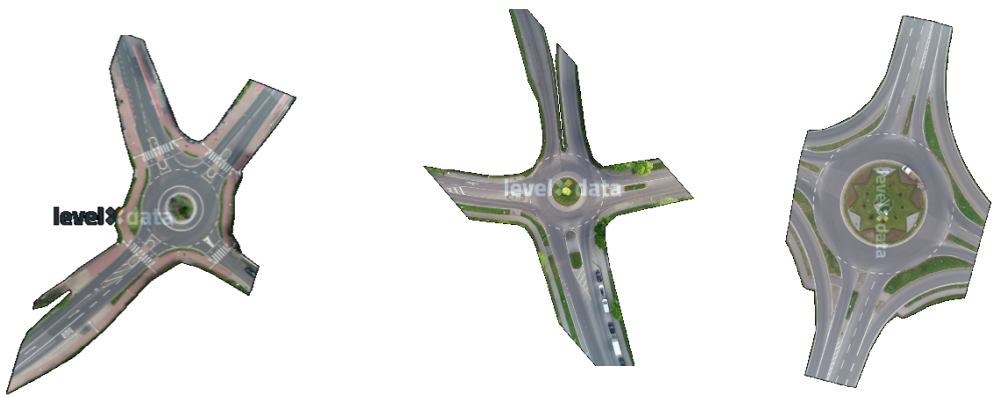

Fig. 1. Images of the three recording sites included in the rounD dataset.

As [6] pointed out, the dataset was created using a pipeline that extracted data from 24 separate recordings of traffic at three different measurement locations in Aachen, Germany. These recordings captured more than 6 hours of video, with most of the recordings being made in the mornings to capture high traffic volume and a lot of interaction. From these recordings, the dataset extracted more than 13,000 road users, including cars, trucks, vans, trailers, buses, pedestrians, bicyclists, and motorcycles. The dataset provides detailed information on driver behavior at roundabouts, including how road users enter and exit the roundabouts, how they interact with other road users, and how they signal their intentions. It also includes information on the characteristics of the roundabouts themselves, such as their size, layout, and the presence of lane markings and pedestrian crossings. Importantly, the dataset contains no recorded collisions, indicating that roundabouts can be safe if drivers follow the correct behavior. The dataset is organized into three recording sites located in and around Aachen, Germany. These sites include a four-armed roundabout that connects a highway with Aachen, a roundabout in an urban area of Aachen, and a four-arm roundabout in a suburb of Aachen. Each site has its own unique characteristics, including varying traffic volume, lane markings, and pedestrian crossings. The rounD dataset is a valuable resource for researchers and policymakers seeking to improve roundabout design and operation. By providing detailed information on driver behavior and roundabout characteristics, the dataset can inform efforts to enhance safety, traffic flow, and efficiency at roundabouts.

**Methods**

To classify driving behavior at roundabouts, a previously developed framework from a study on signalized intersections and another study on work zones was utilized [16], [17]. The framework's main components were utilized, but the study focused on different road infrastructure, namely roundabout. The first step involved extracting features from each driver's trajectory data using volatility measures, shown in Table 1. Volatility measures are significant safety parameters for identifying driver behavior and have been used in many studies [16]–[18]. A higher value of volatility measures implies the driver is more unstable and riskier, and hence more aggressive [16]–[18]. Thirteen different volatility measures were used. Next, these extracted features were utilized as input for an unsupervised machine learning algorithm to cluster each driver's behavior in the work zone. The K-means algorithm was used in this study, which had been successful in previous studies [16], [17].

Table 1. Volatility measures used as inputs to the unsupervised machine learning algorithm.

| Volatility Measure | Description | Equation |
|---|---|---|
| $DV_1$ | Standard deviation of speed | $\sqrt{\frac{\sum_{i=1}^{N}(V_i-\overline{V})^2}{N}}$ |
| $DV_2$ | Standard deviation of longitudinal deceleration or acceleration | $\sqrt{\frac{\sum_{i=1}^{N}(AD_{long_i}-\overline{AD_{long}})^2}{N}}$ |
| $DV_3$ | Coefficient of variation of speed | $100 \times \frac{\sqrt{\frac{\sum_{i=1}^{N}(V_i-V)^2}{N}}}{\overline{V}}$ |
| $DV_4$ | Coefficient of variation of longitudinal acceleration | $100 \times \frac{\sqrt{\frac{\sum_{i=1}^{N}(A_{long_i}-A_{long})^2}{N}}}{\overline{A_{long}}}$ |
| $DV_5$ | Coefficient of variation of longitudinal deceleration | $100 \times \frac{\sqrt{\frac{\sum_{i=1}^{N}(D_{long_i}-D_{long})^2}{N}}}{\overline{D_{long}}}$ |
| $DV_6$ | Mean absolute deviation of speed | $\frac{\sum_{i=1}^{N}|V_i-\overline{V}|}{N}$ |
| $DV_7$ | Mean absolute deviation of longitudinal acceleration | $\frac{\sum_{i=1}^{N}|A_{long_i}-\overline{A_{long}}|}{N}$ |
| $DV_8$ | Quantile coefficient of variation of normalised speed | $100 \times \frac{Q_{V_3}-Q_{V_1}}{Q_{V_3}+Q_{V_1}}$, where $Q_1$ and $Q_3$ are the sample $25^{th}$ and $75^{th}$ percentiles. |
| $DV_9$ | Quantile coefficient of variation of longitudinal acceleration | $100 \times \frac{Q_{A_{long_3}}-Q_{A_{long_1}}}{Q_{A_{long_3}}+Q_{A_{long_1}}}$ |
| $DV_{10}$ | Quantile coefficient of variation of longitudinal deceleration | $100 \times \frac{Q_{D_{long_3}}-Q_{D_{long_1}}}{Q_{D_{long_3}}+Q_{D_{long_1}}}$ |
| $DV_{11}$ | Percentage of time the mean normalised speed exceeds the mean plus two standard deviations | $100 \times \frac{\sum_{i=1}^{N}(V_i \geq \overline{V}+2*\alpha)}{N}, \alpha = DV_1$ |
| $DV_{12}$ | Percentage of time the mean of longitudinal acceleration exceeds the mean plus two standard deviations | $100 \times \frac{\sum_{i=1}^{N}(A_{long_i} \geq \overline{A_{long}}+2*\alpha)}{N}, \alpha = DV_2$ |

| $DV_{13}$ | Percentage of time the mean longitudinal deceleration exceeds the mean plus two standard deviations | $100 \times \frac{\sum_{i=1}^{N}\left(D_{long_i} \geq \overline{D_{long}} + 2*\alpha\right)}{N}, \alpha = DV_2$ |
|---|---|---|

K-means algorithm, which was used in this study, is a popular unsupervised machine learning algorithm used for clustering and data partitioning. The algorithm partitions a dataset into K different clusters based on the similarity between data points. The number of clusters, K, is specified beforehand by the user. The algorithm iteratively assigns data points to the nearest cluster based on the Euclidean distance between the data point and the centroid of the cluster. The centroid is recalculated after each iteration based on the mean value of all the data points in the cluster. The algorithm continues to iterate until there is no significant change in the assignment of data points to clusters. The K-means algorithm is widely used in various applications, including customer segmentation, image compression, and anomaly detection.

## 4. Analysis and Results

K-means algorithm was used to cluster driving behavior based on the volatility measures by finding their centroid points. A centroid point is the average of all the data points in the cluster. By iteratively assessing the Euclidean distance between each point in the dataset, each one can be assigned to a cluster. The centroid points are initially assigned randomly and will change each time as the process is carried out. K-means is commonly used in cluster analysis and has been proven to be useful in such cases [16], [17], [19].

The Elbow method is usually used to find the optimal number of clusters for the K-means algorithm using the thirteen volatility measures. The optimal number of clusters can be chosen as two or three clusters as it produces a relatively low total distortion and can be physically interpreted. We tested both cases to compare their results. Each cluster was labeled as 1, 2, or 3 and each of them indicates a classified driving behavior. Determining which driving behavior is assigned to a cluster was based on the mean values of classification features. After performing K-means using all possible features for $k = 2$ and $k = 3$, results are presented in Table 2.

Table 2. The scaled cluster centres at roundabouts for $k = 2$ and $k = 3$.

| Volatility Measures | $k = 3$ | | | $k = 2$ | |
|---|---|---|---|---|---|
| | Cluster 1 (Conservative) | Cluster 2 (Normal) | Cluster 3 (Aggressive) | Cluster 1 (Conservative) | Cluster 2 (Normal) |
| $DV_1$ | 3.03 | 2.92 | 5.76 | 2.92 | 3.03 |
| $DV_2$ | 0.74 | 0.73 | 0.63 | 0.73 | 0.74 |
| $DV_3$ | 116.80 | -103.26 | -6152.43 | -107.88 | 116.80 |
| $DV_4$ | 71.82 | 69.06 | 75.97 | 69.07 | 71.82 |
| $DV_5$ | -68.78 | -67.45 | -79.26 | -67.46 | -68.78 |
| $DV_6$ | 2.64 | 2.54 | 4.95 | 2.54 | 2.65 |
| $DV_7$ | 0.408 | 0.38 | 0.44 | 0.38 | 0.41 |
| $DV_8$ | 93.56 | -83.72 | -222.80 | -83.82 | 93.56 |
| $DV_9$ | 54.03 | 50.89 | 65.79 | 50.89 | 54.03 |
| $DV_{10}$ | -53.44 | -52.75 | -41.83 | -52.74 | -53.44 |
| $DV_{11}$ | 361.17 | -98.33 | 243.79 | -98.06 | 361.72 |
| $DV_{12}$ | 52.33 | 51.53 | 51.65 | 51.53 | 52.33 |
| $DV_{13}$ | -10.36 | -10.49 | -8.69 | -10.48 | -10.36 |
| Sample Size | 6967 | 6535 | 5 | 6967 | 6540 |

According to the clustering results, we found that most of the drivers proceeding through a roundabout can be mostly classified into two driving styles: conservative and normal. This is due to many factors. First, it is usually drivers' responsibility when approaching a roundabout to yield to the traffic already in the roundabout (or pedestrians and bicyclists if there is a crosswalk or a bike lane) and only merge when there is a safe gap in the traffic to do so. Thus, roundabouts are generally considered to reduce traffic crashes, because traffic speeds in roundabouts are relatively lower than in other signalized and unsignalized intersections. It is also considered that there are less conflict points in roundabouts than any other road infrastructure.

The other goal of this study is to further investigate the behavior of the drivers that interacted with a VRU (i.e., a pedestrian or a bike). There were 113 VRUs and 13,507 vehicles. Out of those, about 3,681 drivers have interacted with a pedestrian or a bike while proceeding through the roundabouts. We found the interaction between the VRUs and drivers of the other vehicles near the roundabouts by matching the position during a specified interval of time. Results of clustering drivers that interacted with VRUs compared with drivers that did not interact with VRUs is shown in Table 3.

Table 3. Results of clustering all drivers, drivers with interaction, and with no interaction.

| Driving Style | All drivers | | Drivers with no interaction | | Drivers with interaction | |
|---|---|---|---|---|---|---|
| | Number | Percentage | Number | Percentage | Number | Percentage |
| Conservative | 6967 | 51.58% | 4125 | 42.06% | 2842 | 77.21% |
| Normal | 6535 | 48.38% | 5692 | 57.93% | 843 | 22.68% |
| Aggressive | 5 | 0.04% | 1 | 0.01% | 4 | 0.11% |
| Total | 13507 | | 9826 | | 3681 | |

Results showed that the percentage of conservative drivers among whom interacted with VRUs (about 77.21%) was significantly higher than conservative drivers that did not interact or the percentage of all conservative drivers. We also found that most of the drivers that were identified as aggressive (about 4 out of 5) were also from the drivers who interacted with VRUs as they proceeded the roundabouts. This means that although drivers tend to slow down as they approach roundabouts, they tend to behave abnormally (conservatively or aggressively) as they interact with VRUs at the roads. This raises a concern as conservative behavior might also increase the risk of crashes, especially rear-end ones. Comparing the effect of the surrounding environment and their interaction with other modes of transportation is crucial for policymakers to determine the effective and suitable safety countermeasures in multimodal intersections.

The interaction of vehicles and pedestrians at roundabouts can have a significant impact on traffic flow. Pedestrians and cyclists are vulnerable road users who require special attention from drivers, especially at roundabouts where the flow of traffic can be complex and unpredictable. Drivers need to be aware of the presence of pedestrians and cyclists and yield to them as necessary. The presence of pedestrians and cyclists at roundabouts can cause delays and disruptions to the flow of traffic. Pedestrians may cross the roundabout at unmarked or marked crosswalks, causing drivers to slow down or stop. Cyclists may also use the roundabout, and drivers need to be aware of their movements and adjust their speed accordingly. In addition, the interaction between pedestrians and cyclists can also impact traffic flow, as they may cross each other's paths and cause further delays. Efforts to improve the interaction of vehicles and pedestrians at roundabouts can help to improve traffic flow. For example, providing marked crosswalks and bicycle lanes can help to better define the paths of pedestrians and cyclists, reducing the likelihood of conflicts with vehicles. Improved signage and education for drivers can also help to increase awareness of the presence of pedestrians and cyclists at roundabouts, reducing the likelihood of accidents and delays.

Moreover, the interaction of vehicles and pedestrians can impact driving behavior in a number of ways. When drivers encounter pedestrians, they may need to slow down or stop, which can lead to changes in their speed and acceleration. This can affect the flow of traffic and cause congestion, particularly in areas with heavy pedestrian traffic. Additionally, drivers may need to be more attentive to their surroundings when pedestrians are present, which can lead to changes in their driving behavior such as increased lane changes, braking, or steering. Pedestrians can also impact the behavior of other road users, such as bicyclists or motorcyclists, who may need to take extra precautions to avoid collisions with pedestrians. Overall, the interaction of vehicles and pedestrians can create complex and dynamic traffic scenarios that require careful attention and awareness from all road users.

## 5. Conclusion

The issue of identifying driving behavior has emerged as driving habit of drivers has a significant impact on road safety. Roundabouts are particularly intriguing due to the high level of user engagement that a driver or an automated vehicle must consider while proceeding through. Using data-driven unsupervised machine learning to categorize driving behavior at roundabouts, we extracted the volatility measures to classify driving behavior and investigate the effect of interaction between drivers and pedestrians or cyclists.

We found that most of the drivers proceeding through a roundabout can be mostly classified into two driving styles: conservative and normal. Roundabouts are generally considered to reduce traffic crashes, because traffic speeds in roundabouts are relatively lower than in other signalized and unsignalized intersections. It is also considered that there are less conflict points in roundabouts than any other road infrastructure. Results also showed that the percentage of conservative drivers among whom interacted with VRUs (about 77.21%) was significantly higher than conservative drivers that did not interact or the percentage of all conservative drivers. Drivers tend to behave abnormally as they interact with VRUs at the roads. This raises a concern as conservative behavior might also increase the risk of crashes, especially rear-end ones. Driving behavior is considered one of the most critical criteria besides having a high uncertainty level in traffic safety studies. Results of this study could be helpful in improving the safety of roads by allowing policymakers to determine the effective and suitable safety countermeasures. Results will also be beneficial for the Advanced Driver Assistance System (ADAS) as the technology is being deployed in a mixed traffic environment.